\documentclass[conference]{IEEEtran}
\IEEEoverridecommandlockouts
\usepackage{cite}
\usepackage{graphicx}
\usepackage{textcomp}
\usepackage{gensymb}
\usepackage[bookmarks=false]{hyperref}
\usepackage{subfig}
\usepackage{tikz}
\usepackage{xcolor}
\usepackage{multirow,booktabs}
\usepackage{float}
\usepackage{amsmath,amssymb,amsfonts}
\usepackage{algorithm}
\usepackage{colortbl}
\usepackage{tabularx}
\usepackage{siunitx}
\usepackage{listing}
\usepackage{color,soul}

\usepackage[utf8]{inputenc}
\usepackage{algcompatible,amssymb}

\def\BibTeX{{\rm B\kern-.05em{\sc i\kern-.025em b}\kern-.08em
    T\kern-.1667em\lower.7ex\hbox{E}\kern-.125emX}}

\newcommand*\circled[1]{\tikz[baseline=(char.base)]{
            \node[shape=circle,fill,inner sep=1pt] (char) {\textcolor{white}{\textbf{#1}}};}}
    
\newcommand{\youtubelink}{\url{https://youtu.be/JKY03NV3C2s}}
\newcommand{\gitlink}{\url{https://github.com/pulp-platform/pulp-dronet}}

\soulregister\cite7
\soulregister\ref7
\soulregister\SI7
\soulregister\circled7

\begin{document}

\title{An Open Source and Open Hardware\\Deep Learning-powered Visual Navigation Engine for Autonomous Nano-UAVs}

\author{\IEEEauthorblockN{Daniele Palossi\IEEEauthorrefmark{1}, Francesco Conti\IEEEauthorrefmark{1}\IEEEauthorrefmark{2}, and Luca Benini\IEEEauthorrefmark{1}\IEEEauthorrefmark{2}}

\IEEEauthorblockA{\IEEEauthorrefmark{1}Integrated System Laboratory - ETH Z\"urich, Switzerland}

\IEEEauthorblockA{\IEEEauthorrefmark{2} Department of Electrical, Electronic and Information Engineering - University of Bologna, Italy}

Email: name.surname@iis.ee.ethz.ch
}

\maketitle

\begin{abstract}
Nano-size unmanned aerial vehicles (UAVs), with few centimeters of diameter and sub-10 Watts of total power budget, have so far been considered incapable of running sophisticated visual-based autonomous navigation software without external aid from base-stations, ad-hoc local positioning infrastructure, and powerful external computation servers.
In this work, we present what is, to the best of our knowledge, the first \SI{27}{\gram} nano-UAV system able to run aboard an end-to-end, closed-loop visual pipeline for autonomous navigation based on a state-of-the-art deep-learning algorithm, built upon the open-source CrazyFlie 2.0 nano-quadrotor.
Our visual navigation engine is enabled by the combination of an ultra-low power computing device (the GAP8 system-on-chip) with a novel methodology for the deployment of deep convolutional neural networks (CNNs).
We enable onboard real-time execution of a state-of-the-art deep CNN at up to \SI{18}{\hertz}.
Field experiments demonstrate that the system's high responsiveness prevents collisions with unexpected dynamic obstacles up to a flight speed of \SI{1.5}{\meter/\second}.
In addition, we also demonstrate the capability of our visual navigation engine of fully autonomous indoor navigation on a \SI{113}{\meter} previously unseen path.
To share our key findings with the embedded and robotics communities and foster further developments in autonomous nano-UAVs, we publicly release all our code, datasets, and trained networks.
\end{abstract}


\section*{Supplementary Material}
Supplementary video at: \youtubelink.
The project's code, datasets and trained models are available at: \gitlink.

\section{Introduction}\label{sec:intro}

Nowadays we are witnessing a proliferation of industrial and research works in the field of autonomous small-size unmanned aerial vehicles (UAVs)~\cite{dunkley14iros,40gSTM32OpticFlow,palossi2017ultra,briod2013optic,7080923,palossi2017target,kang2019generalization,7487496}.
This considerable effort can be easily explained by the potential applications that would greatly benefit from intelligent miniature robots.
Many of these works, despite they refer to them self as ``autonomous'', are actually ``automatic'' but not independent from some external ad-hoc signal/computation (e.g., GPS, RFID signals, and ground-stations).
We belive that achieving full independence is the key condition to be truly autonomous.

\begin{figure}[t]
\centerline{\includegraphics[width=\columnwidth]{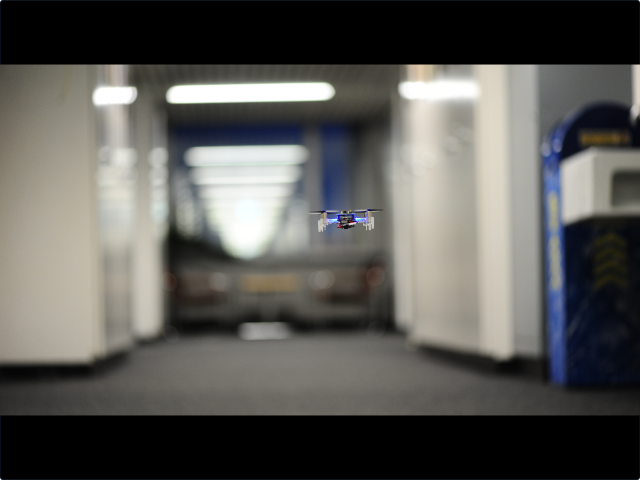}}
\caption{Our prototype deployed on the filed. It is based on the \textit{Crazyflie 2.0} nano-quadrotor extended with our \textit{PULP-Shield}. The system can run the \textit{DroNet}~\cite{dronet} CNN for autonomous visual navigation up to \SI{18}{\hertz} using only onboard resources.}
\label{fig:prototype}
\end{figure}

Autonomous pocket-size drones can be particularly versatile and useful, acting as sensor nodes that acquire information, process and understand it, and use it to interact with the environment and with each other.
The ``ultimate'' will be capable of autonomously navigating the environment and, at the same time, sensing, analyzing, and understanding it~\cite{iotUAV_survey}.
In the context of wireless sensor networks (WSNs), such a miniaturized robotic-helpers can collect the data from a local WSN and bridge them towards the external world.
Moreover, a swarm of such intelligent and ubiquitous nano-drones can quickly build a remote sensing network in an emergency context, where their small size enables inexpensive and fast deployment on the field, as well as reaching location inaccessible for human operators or standard-size drones.
The tiny form-factor of nano-drones (i.e., featuring only few centimeters in diameter and few tens of grams in weight) is ideal both for indoor applications where they should safely operate near humans and for highly-populated urban areas, where they can exploit complementary sense-act capabilities to interact with the surroundings (e.g., smart-building, smart-cities, etc.).

To enable such ambitious scenarios many challenging problems must be addressed and solved.
Nano-scale commercial off-the-shelf (COTS) quadrotors still lack a meaningful level of autonomy, contrary to their larger counterparts~\cite{dronet, Lin_2017, loianno2018special}, since their tiny power envelopes and limited payload do not allow to host onboard adequate computing power for sophisticated workloads.
Of the total power available on a UAV (listed in Table~\ref{tab:taxonomy} for four classes of vehicles), Wood~et~al.~\cite{Wood2017} estimate that only up to 5\% is available for onboard computation, and payloads of maximum $\sim$25\% of the total mass can be allotted to the electronics.

\begin{table}[t]
\renewcommand{\arraystretch}{1.5}
\caption{Rotorcraft UAVs taxonomy by vehicle class-size.}
\label{tab:taxonomy}
\centering
\footnotesize
\resizebox{\columnwidth}{!}{
\begin{tabular}{|c|c|c|c|}
\hline
Vehicle Class & $\oslash$ : Weight [cm:kg] & Power [W] & Onboard Device\\
\hline
\text{\textit{std-size} \cite{dronet}}		& $\sim$ 50 : $\geq$ 1		& $\geq$ 100	& Desktop\\ 
\hline
\text{\textit{micro-size} \cite{conroy2009}}	    & $\sim$ 25 : $\sim$ 0.5 	& $\sim$ 50		& Embedded\\ 
\hline
\text{\textit{nano-size} \cite{40gSTM32OpticFlow}}	& $\sim$ 10 : $\sim$ 0.01 	& $\sim$ 5		& MCU\\ 
\hline
\text{\textit{pico-size} \cite{Wood2017}} 	        & $\sim$ 2 : $\leq$ 0.001	& $\sim$ 0.1 	& ULP\\ 
\hline
\end{tabular}
}
\end{table}

The traditional approach to autonomous navigation of a UAV is the so-called \textit{localization-mapping-planning} cycle, which consists of estimating the robot motion using either off-board (e.g., GPS) or onboard sensors (e.g., visual-inertial sensors), building a local 3D map of the environment, and planning a safe trajectory through it~\cite{loianno2018special}. 
These methods, however, are very expensive for computationally-constrained platforms.
Recent results have shown that much lighter algorithms, based on convolutional neural networks (CNNs), are sufficient for enabling basic reactive navigation of small drones, even without a map of the environment~\cite{dronet, giusti2016machine}. 
However, their computational and power needs are unfortunately still above the allotted budget of current navigation engines of nano-drones, which are based on simple, low-power microcontroller units (MCUs).

In this work, we introduce several improvements over the state of the art of nano-scale UAVs.
First, we introduce the design of a low-power visual navigation module, the \textit{PULP-Shield}, featuring the high-efficiency \textit{GreenWaves Technologies GAP8 SoC}, a ULP camera and Flash/DRAM memory, compatible with the \textit{CrazyFlie} 2.0 nano-UAV.
The full system is shown in the field in Figure~\ref{fig:prototype}.
We propose a methodology for embedding the CNN-based \textit{DroNet}\cite{dronet} visual navigation algorithm, originally deployed on standard-sized UAVs with off-board computation, in a nano-UAV with fully onboard computation.
We demonstrate how this methodology yields comparable quality-of-results (QoR) with respect to the baseline, within a scalable power budget of \SI{64}{\milli\watt} at 6 frames per second (fps), up to \SI{272}{\milli\watt} at 18 fps.

We prove in the field the efficacy of this methodology by presenting a closed-loop fully functional demonstrator in the supplementary video material, showing autonomous navigation on \textit{i}) a \SI{113}{\meter} previously unseen indoor environment and \textit{ii}) collision robustness against the appearance of a sudden obstacle at a distance of \SI{2}{\meter} while flying at \SI{1.5}{\meter/\second}.
To the best of our knowledge, our design is the first to enable such complex functionality in the field on a nano-UAV consuming $<$\SI{100}{\milli\watt} for electronics.
To foster further research on this field, we release the PULP-Shield design and all code running on GAP8, as well as datasets and trained networks, as publicly available under liberal open-source licenses.

\section{Related Work}\label{sec:related}

The traditional approach to the navigation of nano-drones requires to offload the computation to a remote  base-station~\cite{dunkley14iros,7080923,kang2019generalization}, demanding high-frequency video streaming, which lowers reliability and imposes constraints on maximum distance, introduces control latency and is poorly scalable.
On the other hand, COTS nano-size quadrotors, like the \textit{Bitcraze Crazyflie 2.0} or the \textit{Walkera QR LadyBug}, usually make use of very simple computing devices such as single-core microcontroller units (MCUs) like the \textit{ST Microelectronics} STM32F4~\cite{40gSTM32OpticFlow,dunkley14iros,7487496}.
Autonomous flying capabilities achievable on these platforms are, to the date, very limited.
In~\cite{40gSTM32OpticFlow} the proposed obstacle avoidance functionality requires favorable flight conditions (e.g., low flight speed of \SI{0.3}{\meter/\second}).
The solutions proposed in~\cite{briod2013optic,palossi2017ultra} are limited to hovering and do not reach the accuracy of computationally expensive techniques leveraged by powerful standard-size UAVs.
\cite{7487496} addresses only state estimation -- a basic building block of autonomous UAVs, but far from being the only required functionality.

An emerging trend in the evolution of autonomous navigation systems is the design and development of application-specific integrated circuit (ASIC) addressing specific navigation tasks~\cite{navion,CNN-SLAM}.
ASICs deliver levels of performance and energy efficiency for the specific tasks addressed that cannot be achieved by typical nano-UAV computing platforms for workloads such as visual odometry~\cite{navion} or simultaneous localization and mapping (SLAM)~\cite{CNN-SLAM}.
However, ASICs only accelerate a part of the overall functionality, requiring pairing with additional circuits for complementary onboard computation as well as for interacting with the drone's sensors.
Moreover, to date, systems based on these ASICs have not yet been demonstrated on board a real-life flying nano-UAV. 

In this work, we demonstrate a sophisticated visual navigation engine that is entirely based on a general-purpose parallel, ultra-low power (PULP) computing platform, and works in closed-loop in the field within the power envelope and payload of nano-scale UAVs ($\sim$\SI{0.2}{\watt} and $\sim$\SI{15}{\gram}, respectively).

\section{Implementation}\label{sec:impl}

This section gives insight on DroNet, the key driving algorithm used by our visual navigation engine, on the hardware platform utilized in this work (the GAP8 SoC), and on how the algorithm was modified to fit within the constrained hardware platform while keeping the same original accuracy and the performance.

\subsection{The algorithm: DroNet}

\begin{figure}[t]
	\centerline{\includegraphics[width=\columnwidth]{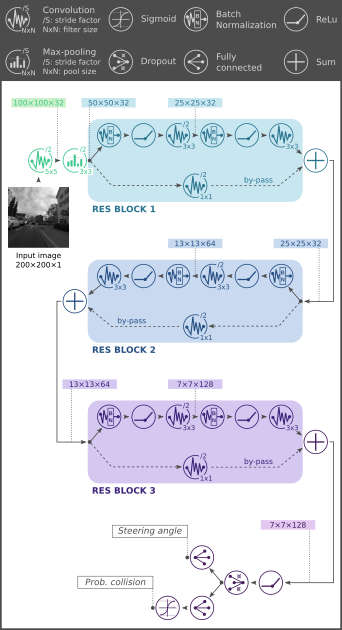}}
	\caption{\textit{DroNet}~\cite{dronet} CNN topology.}
	\label{fig:dronet-archi}
\end{figure}

The key driver for our proposed autonomous visual navigation engine is \textit{DroNet}: an algorithm proposed initially by Loquercio~et~al.~\cite{dronet} based on a convolutional neural network (CNN) whose topology is inspired on ResNet~\cite{he2016deep}.
The original DroNet was deployed on top of a commercial standard-size UAV streaming camera frames to an external laptop.
DroNet is trained to convert an unprocessed input image from a camera into two high-level pieces of information: \textit{i}) an estimation of the probability of \textit{collision} with an obstacle, which in turn can be used to determine the forward target velocity of the UAV; \textit{ii}) the desired \textit{steering} direction, following visual cues from the camera such as the presence of obstacles, white lines on the floor or in the street, etc.

Figure~\ref{fig:dronet-archi} reports the full topology of DroNet, which is shared between the two steering and collision tasks up to the penultimate layer.
To train the network\footnote{Following~\cite{dronet}, the steering and collision tasks were associated to mean squared error (MSE) and binary cross-entropy (BCE) losses, respectively. The Adam optimizer was used, with starting learning rate of $1^{-3}$ and learning rate decay per epoch equal to $1^{-5}$.
We refer to Loquercio~et~al.~\cite{dronet} for further details on the training methodology.} two openly available datasets were used -- \textit{Udacity}\footnote{https://www.udacity.com/self-driving-car}, a dataset designed to train self-driving cars, for the steering task, and the \textit{Z\"{u}rich bicycle} dataset\footnote{http://rpg.ifi.uzh.ch/dronet.html} for the collision task.

At inference time, the steering direction $\theta_{steer}$ and collision $P_{coll}$ outputs of the network are connected to the UAV control, influencing the target yaw rate $\omega_{yaw,target}$ and the target forward velocity $v_{x,target}$ through a simple low-pass filtering scheme:

\begin{align}
    v_{x,target}[t] &= \alpha \cdot v_{\max} \cdot (1 - P_{coll}[t]) \\
                  &+ (1-\alpha) \cdot v_{x,target}[t-1]\notag\\
    \omega_{yaw,target}[t] &= \beta \cdot \theta_{steer}[t] \\
                         &+ (1-\beta) \cdot \omega_{yaw,target}[t-1]\notag
\end{align}
where the parameters have default values $\alpha=0.3$ and $\beta=0.5$.

\subsection{The platform: GAP8 SoC}

While commercial off-the-shelf microcontrollers used in the most common nano-UAV platforms have acceptable computing capabilities of their own, these could not be enough to achieve autonomous flight functionality, which requires workloads in the order of 100 million -- 10 billion operations per second~\cite{dronet}.
Moreover, these microcontrollers are typically tasked with many computationally simple but highly critical real-time tasks to estimate the current kinematic state of the UAV, predict its motion and control the actuators.
To avoid tampering with this mechanism, we chose to execute our visual navigation engine on a different platform than the central nano-UAV microcontroller, acting as a specialized accelerator~\cite{contiDATE16} based on the GreenWaves Technologies GAP8 system-on-chip (SoC).

\begin{figure}[h]
	\centering
	\includegraphics[width=\columnwidth]{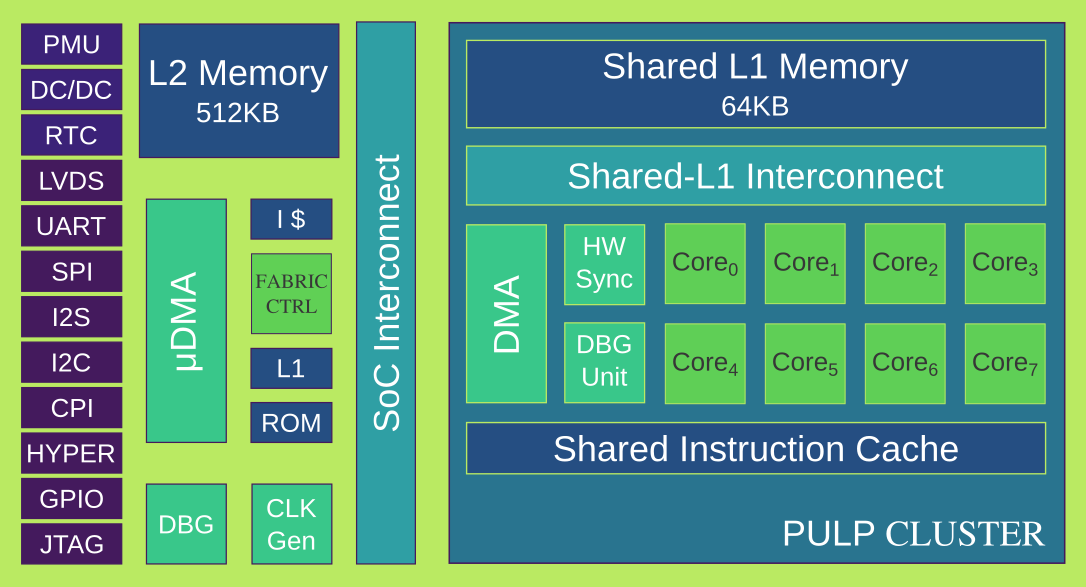}
	\caption{Architecture of the PULP-GAP8 embedded processor.}
	\label{fig:gap8_archi}
\end{figure}

GAP8 is a commercial embedded application processor based on the PULP open source architecture\footnote{http://pulp-platform.org} and the RISC-V open ISA.
Figure~\ref{fig:gap8_archi} shows the architecture of GAP8 in detail.
The GAP8 SoC is organized in two subsystems and power domains, a \textit{fabric controller} (FC) with one RISC-V core acting as an on-SoC microcontroller and a \textit{cluster} (CL) serving as an accelerator with 8 parallel RISC-V cores.
All the cores in the system are identical and support the RV32IMC instruction set with SIMD DSP extensions (e.g., fixed-point dot product) to accelerate linear algebra and signal processing.

The FC is organized similarly to a microcontroller system, featuring an internal clock generator, \SI{512}{\kilo\byte} of SRAM (L2 memory), a ROM for boot code, and an advanced I/O subsystem ($\mu DMA$) that can be programmed to autonomously move data between a wide set of I/O interfaces (including SPI, UART, I2C, L3 HyperRAM) and the L2 memory without the core's intervention.
The CL is meant to be used to accelerate parallel sections of the application code running on GAP8.
Its 8 RISC-V cores share a single shared cache for instructions and a shared L1 scratchpad memory of \SI{64}{\kilo\byte} for data; movement of data between the latter and the L2 is manually managed by the software running on the cluster using an internal DMA controller.
This enables us to achieve maximum efficiency and utilization on typical parallel kernels with regular, predictable access patterns for data while saving the area overhead of a shared data cache.

\subsection{Optimizations for embedded deployment}

\begin{figure}[t]
	\centerline{\includegraphics[width=0.9\columnwidth]{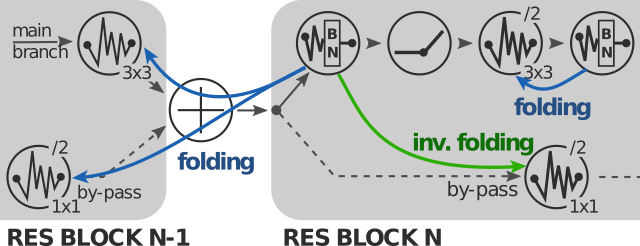}}
	\caption{Batch-normalization layer folding methodology.}
	\label{fig:bn_folding}
\end{figure}

Deploying a CNN algorithm developed in a high-level framework (such as TensorFlow in the case of DroNet) to a low-power application processor such as GAP8 involves several challenges, connected with the constraints imposed by the limited available resources.
First, the navigation algorithm must be able to execute the main workload ($\sim$ 41 million of multiply-accumulate operations for one inference pass-through of DroNet) at a frame rate sufficient to achieve satisfactory closed-loop performance in control.
Furthermore, while the embedded processor typically uses a lower precision to represent data and a lower resolution input camera, the quality-of-results must remain similar to the one of the original algorithm.
These constraints impose significant modifications to the original algorithm that in the case of DroNet can be grouped in two main categories.

\subsubsection{Dataset fine-tuning \& network quantization}

To improve the generalization capabilities of the original DroNet~\cite{Razavian2014transfer} with respect to the lower-quality images coming from the embedded camera, we collected an extension for the collision dataset using directly the camera available in the final platform: a grayscale QVGA-resolution HiMax.
We collected 1122 new images for training and 228 for test/validation, which we compounded with the openly available collision dataset.
We also replaced $3\times 3$ pooling layers with $2\times 2$ ones, which yield the same overall functionality (i.e., the reduction of the spatial size of feature maps in the CNN) while being smaller and generally easier to implement as each input pixel is projected to a single output one.
Finally, to adapt the network to execution on a low-power platform without support for floating-point numbers, we switched to fixed-point data representation.
Specifically, by analyzing the dynamic range of intermediate feature maps in the original DroNet, we found that a precision of $2^{-11}$ and a range $\pm 16$ was adequate to represent activation data after batch normalization (BN) layers.
Then, we replaced all activation ReLU layers with quantization-aware equivalents~\cite{HubaraQuantizedNeuralNetworks2016} using a 16-bit Q5.11 fixed-point format.
The entire network was retrained from scratch using the same framework of the original DroNet.

\subsubsection{Batch-norm folding}

During training, batch-normalization (BN) layers are essential to keep the dynamic range of feature map activations in check (hence helping with their quantization) and to regularize the learning process, which achieves a far better results in terms of generalization than an equivalent network, particularly for what concerns the regression task of computing the desired steering.
However, during inference, the BN layers are linear and can be merged with the preceding convolutional layer by \textit{folding} it inside its weights $W$ and biases $b$.
If $\gamma$, $\beta$, $\sigma$, and $\mu$ are the normalization parameters, then:

\begin{align}
	\mathrm{BN}\big(\mathbf{W}\star\mathbf{x} + \mathbf{b}\big) &= \gamma/\sigma \cdot \left(\mathbf{W}\star\mathbf{x}+\mathbf{b} -\mu\right) + \beta \notag \\
	&= \left(\gamma/\sigma \cdot \mathbf{W}\right) \star \mathbf{x} + \Big(\beta+\gamma/\sigma \left(\mathbf{b} -\mu\right)\Big) \notag \\
	&\doteq \mathbf{W'} \star \mathbf{x} + \mathbf{b'} \label{eq:bn_folding}
\end{align}

In DroNet, the input of each RES block is normalized in the main branch, but non-normalized in the by-pass branch, making the direct application of Equation~\ref{eq:bn_folding} more difficult, as it is not possible to directly apply it to the convolution preceding those operations.
Therefore, we proceeded as follows: for each RES block, we first apply the folding ``as if'' the input of the entire RES block was normalized by using Equation~\ref{eq:bn_folding}.
This means that each BN is folded into the previous convolution layer, e.g., for RES block 1, in the initial convolutional layer of DroNet and in the first one of the main branch.
Second, we apply \textit{inverse folding} on the by-pass convolutional layer, to counteract the folding of BNs on its inputs:

\begin{align}
    \mathrm{BN}^{-1}&\big(\mathbf{W'}\star\mathbf{x} + \mathbf{b'}\big) \doteq \mathbf{W''}\star\mathbf{x} + \mathbf{b''} \\
    \mathbf{W''} &\doteq \sigma/\gamma \cdot \mathbf{W'} \notag \\
    \mathbf{b''} &\doteq \mathbf{b'} + \sum_{ic}\Big(\mu \cdot \sum_{fs}\mathbf{W'}\Big) - \sum_{ic}\Big(\beta \cdot \sigma/\gamma \cdot \sum_{fs}\mathbf{W'}\Big) \notag
\end{align}

where $\sum_{ic}$ and $\sum_{fs}$ indicate marginalization along the input channels dimension and along the filter's spatial dimensions, respectively. 
We apply this operation sequentially to each RES block as exemplified in Figure~\ref{fig:bn_folding}.
After this operation, the BN layers can be effectively removed from the network as other layers absorb their effects.
Finally, the new weights and bias values can be quantized according to their range requirements.
In the final DroNet deployment, we quantize weights for all layers at Q2.14, except for the first bypass layer, which uses Q9.7.

\section{The PULP-Shield}\label{sec:proto}

\begin{figure*}[t]
	\centering
	\includegraphics[width=\textwidth]{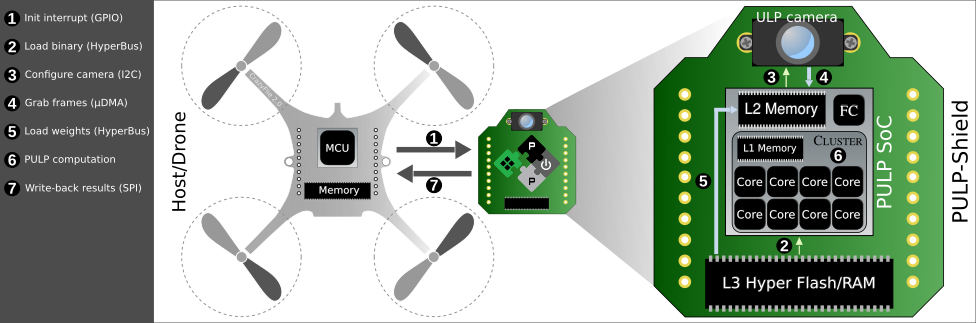}
	\caption{Interaction between the \textit{PULP-Shield} and the \textit{CrazyFlie 2.0} nano-drone.}
	\label{fig:pulp-shield-model}
\end{figure*}

Our visual navigation engine is embodied, on its hardware side, in the so-called \textit{PULP-Shield}: a lightweight, modular and configurable printed circuit board (PCB) with a highly optimized layout.
We designed the PULP-Shield to be compatible/pluggable to the \textit{Crazyflie 2.0} (CF) nano-quadrotor\footnote{https://www.bitcraze.io/crazyflie-2}.
The CF has been chosen due to its reduced size (i.e., \SI{27}{\gram} of weight and \SI{10}{\centi\meter} of diameter), its open-source and open-hardware philosophy, and the availability of extra payload (up to \SI{15}{\gram}).
The PULP-shield features a PULP-based GAP8 SoC, two Cypress \textit{HyperBus Memories}\footnote{http://www.cypress.com/products/hyperbus-memory} enabling flexible configuration and an ultra-low-power gray-scale \textit{HiMax}\footnote{http://www.himax.com.tw/products/cmos-image-sensor/image-sensors} QVGA CMOS image sensor that communicates via the parallel camera interface (PCI) protocol.
On the two BGA memory slots we mounted a \SI{64}{\mega\bit} \textit{HyperRAM} (DRAM) chip and a \SI{128}{\mega\bit} \textit{HyperFlash} memory, embodying the system L3 and the external storage, respectively.

\begin{figure}[h]
	\centering
	\includegraphics[width=\columnwidth]{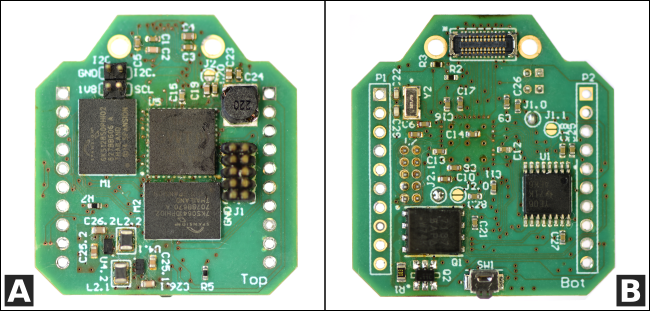}
	\caption{The \textit{PULP-Shield} pluggable PCB.}
	\label{fig:pulp-shield-schematic}
\end{figure}

Two mounting holes, on the side of the camera connector, allow to plug a 3D-printed camera holder that can be set either in front-looking or down-looking mode, accounting for the most common visual sensors layouts and enabling a large variety of tasks like obstacle avoidance~\cite{dronet} and visual state estimation~\cite{palossi2017ultra}, respectively.
On the shield there are also a JTAG connector for debug purposes and an external I2C plug for future development.
Two headers, located on both sides of the PCB, grant a steady physical connection with the drone and at the same time they bring the shield power supply and allow communication with the CF's main MCU (i.e., \textit{ST Microelectronics STM32F405}\footnote{http://www.st.com/en/microcontrollers/stm32f405-415.html}) through SPI interface and GPIO signals.
The form factor of our final PULP-Shield prototype, shown in Figure~\ref{fig:pulp-shield-schematic}, is 30$\times$\SI{28}{\milli\meter} and it weighs $\sim$\SI{5}{\gram} (including all components), well below the payload limit imposed by the nano-quadcopter.

The PULP-Shield embodies the \textit{Host-Accelerator} heterogeneous architectural paradigm at the ultra-low power scale~\cite{contiDATE16}, where the CF's MCU offloads the intensive visual navigation workloads to the PULP accelerator.
As reported in Figure~\ref{fig:pulp-shield-model} the interaction starts from the host, which wakes up the accelerator with a GPIO interrupt \circled{1}.
Then, the accelerator fetches from its external HyperFlash storage the binary to be executed \circled{2}.
After the ULP camera is configured via I2C \circled{3} the frames can be transferred to the L2 shared memory through the $\mu$DMA \circled{4} and this can be performed in pipeline with the computation running on the \textsc{cluster} (i.e., in \textit{double buffering} fashion).
All additional data, like the weights used in our CNN, can be loaded from the DRAM/Flash memory \circled{5} and the parallel execution can start on the accelerator \circled{6}.
Once the computation is completed the results are returned to the drone's MCU via SPI \circled{7}.

Even if the PULP-Shield has been developed specifically to fit the CF quadcopter, its basic concept and the functionality it provides are quite general and portable to any drone based on an SPI-equipped MCU.
The system-level architectural template is meant for minimizing data transfers (i.e., exploiting locality of data) and communication overhead between the main MCU and the accelerator -- without depending on the internal microarchitecture of either one.

\section{Experimental Results}\label{sec:results}

In this section we present the experimental evaluation of our visual navigation engine, considering three main metrics: \textit{i}) a QoR comparison with other CNNs for autonomous navigation of UAVs, \textit{ii)} the capability of performing all the required computations within the allowed power budget and \textit{iii)} a quantitative control accuracy evaluation of the closed-loop system when deployed on the field. 
All the results are based on the PULP-Shield configuration presented in Section~\ref{sec:proto}.

\subsection{CNN Evaluation}

To assess the regression performance of our modifications to the original CNN, employing the testing sequence from the Udacity dataset, we present in Table~\ref{tab:dronet_pulp_accuracy} a comparison with the state-of-the-art.
We compare our version of the DroNet network, named \textit{PULP-DroNet}, against a set of other architectures from the literature~\cite{giusti2016machine,he2016deep,xu2017end} and also against the same original DroNet model~\cite{dronet}.
Note that, we report the same accuracy/performance previously presented in~\cite{dronet} for the same reference architectures.
Our regression and classification results are gathered analyzing the testing sequence on the official PULP simulator, that precisely models the behavior of the target architecture executing the same binary deployed on the  PULP-Shield.
Performance results (e.g., processing time) are instead obtained running the PULP-DroNet CNN on the actual hardware.
In Table~\ref{tab:dronet_pulp_accuracy}, explained variance (EVA) and root-mean-square error (RMSE) refer to the regression problem (i.e., steering angle) whereas Accuracy and F1-score are related to the classification problem (i.e., collision probability).

\begin{table*}[t]
\renewcommand{\arraystretch}{1.3}
\caption{Results on regression and classification task.}
\label{tab:dronet_pulp_accuracy}
\centering
\begin{tabular}{c c c c c c c | c c}
\cline{1-9}
\textbf{Model}   &   \textbf{EVA} &   \textbf{RMSE}    &   \textbf{Accuracy}   &   \textbf{F1-score}   &   \textbf{Num. Layers} &   \textbf{Memory [MB]}   &  \textbf{Processing time [fps]}  &   \textbf{Device}\\
\cline{1-9}
Giusti et al.~\cite{giusti2016machine}   &   0.672   &   0.125   &   91.2\%  &   0.823   &   6   &   0.221   &   23  &   Intel Core i7\\
ResNet-50~\cite{he2016deep}   &   0.795   &   0.097   &   96.6\%  &   0.921   &   50   &   99.182   &   7  &   Intel Core i7\\
VGG-16~\cite{xu2017end}   &   0.712   &   0.119   &   92.7\%  &   0.847   &   16   &   28.610   &   12  &   Intel Core i7\\
DroNet~\cite{dronet}   &   0.737   &   0.109   &   95.4\%  &   0.901   &   8   &   1.221   &   20  &   Intel Core i7\\
\textbf{PULP-DroNet (Ours)}  &  0.748   &   0.111   &   95.9\%  &   0.902   &   8   &  0.610   &   18  &   GAP8 SoC\\
\cline{1-9}
\end{tabular}
\end{table*}

From these results, we can observe that our modified design, even though 160 times smaller and running with two orders of magnitude lower power consumption than the best architecture (i.e., ResNet-50~\cite{he2016deep}), maintains a considerable prediction performance while achieving comparable real-time operation (18 frames per second).
Regarding the original DroNet, it is clear that the proposed modifications, like quantization and fixed-point calculation, are not penalizing the overall network's capabilities, quite the opposite.
In fact, both the regression and classification problems benefit from the fine-tuning, highlighting how the generalization of such models depends critically on the quantity and variety of data available for training.

\subsection{Performance and Power Consumption}

\begin{figure}[t]
	\centering
	\includegraphics[width=\columnwidth]{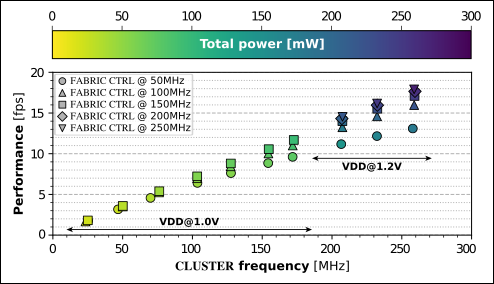}
	\caption{\textit{DroNet} performance in frames per second (fps) in all tested configurations (coloring is proportional to total system power).}
	\label{fig:perfVSfreqVSpower}
\end{figure}

We measured wall-time performance and power consumption by sweeping between several operating modes on GAP8.
We focused on operating at the lowest (\SI{1.0}{\volt}) and highest (\SI{1.2}{\volt}) supported core VDD voltages.
We swept the operating frequency between 50 and \SI{250}{\mega\hertz}, well beyond the GAP8 officially supported configuration\footnote{https://greenwaves-technologies.com/gap8-datasheet}.
In Figure~\ref{fig:perfVSfreqVSpower} we report performance as frame-rate and total power consumption measured on the GAP8 SoC. 
Selecting a VDD operating point of \SI{1.2}{\volt} would increase both power and performance up to \SI{272}{\milli\watt} and \SI{18}{fps}.
We found the SoC to be working correctly @ \SI{1.0}{\volt} for frequencies up to $\sim$\SI{175}{\mega\hertz}; we note that as expected when operating @ \SI{1.0}{\volt} there is a definite advantage in terms of energy efficiency.
We identified the most energy-efficient configuration in VDD@\SI{1.0}{\volt}, FC@\SI{50}{\mega\hertz} and CL@\SI{100}{\mega\hertz}, that is able to deliver up to \SI{6}{fps}, with an energy requirement per frame of \SI{7.1}{\milli\joule}.

\begin{figure}[t]
	\centering
	\includegraphics[width=\columnwidth]{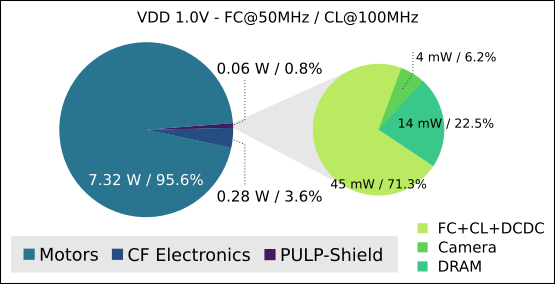}
	\caption{Power envelope break-down of the entire cyber-physical system running at VDD@\SI{1.0}{\volt}, FC@\SI{50}{\mega\hertz}, CL@\SI{100}{\mega\hertz} with \textit{PULP-Shield} zoom-in.}
	\label{fig:break_down}
\end{figure}

In Figure~\ref{fig:break_down}, we report the power break-down for the complete cyber-physical system and for the proposed PULP-Shield. 
Our nano-quadcopter is equipped with a \SI{240}{\milli\ampere\hour} \SI{3.7}{\volt} LiPo battery enabling a flight time of $\sim$7 minutes under standard conditions, which results in an average power consumption of \SI{7.6}{\watt}.
The power consumption of all the electronics aboard the original drone amounts to \SI{277}{\milli\watt} leaving $\sim$\SI{7.3}{\watt} for the 4 rotors.
The electronics consumption is given by the 2 MCUs included in the quadrotor and all the additional devices (e.g., sensors, LEDs, etc.).
In addition to that, introducing the PULP-Shield, we increase the peak power envelope by \SI{64}{\milli\watt} (i.e., 0.8\% of the total) using the most energy-efficient configuration and accounting also for the cost of L3 memory access and the onboard ULP camera.
On the PULP-Shield break-down, visible on the right of  Figure~\ref{fig:break_down}, we consider the worst-case envelope of the HyperRAM operating at full speed only for the time required for L3-L2 data transfers with an average power consumption of \SI{14}{\milli\watt}.
As onboard computation accounts for roughly 5\% of the overall power consumption (propellers, sensors, compute and control, \textit{cfr} Section~\ref{sec:intro}), our PULP-Shield enables the execution of the DroNet network (and potentially more) in all configurations within the given power envelope.

\begin{table}[h]
\renewcommand{\arraystretch}{1.3}
\caption{\textit{CrazyFlie 2.0} (CF) lifetime with and without \textit{PULP-Shield} (both turned off and running \textit{DroNet} at VDD@\SI{1.0}{\volt}, FC@\SI{50}{\mega\hertz}, CL@\SI{100}{\mega\hertz}).}
\label{tab:lifetime}
\centering
\begin{tabular}{|c|c|c|c|c|}
\hline
\multirow{2}{*}{} & \multirow{2}{*}{Original CF} & \multicolumn{2}{c|}{CF + \textit{PULP-Shield}} \\
\cline{3-4}
& & \textit{PULP-Shield} (off) & \textit{PULP-Shield} (on) \\
\hline
Lifetime & $\sim$\SI{440}{\second} & $\sim$\SI{350}{\second} & $\sim$\SI{340}{\second} \\
\hline
\end{tabular}
\end{table}

Finally, we performed an experiment to evaluate the cost in terms of operating lifetime of carrying the physical payload of the PULP-Shield and of executing the DroNet workload.
To ensure a fair measurement, we decoupled the DroNet output from the nano-drone control and statically set it to \textit{hover} (i.e., keep constant position over time) at \SI{0.5}{\meter} from the ground.
We targeted three different configurations: \textit{i}) the original CrazyFlie without any PULP-Shield; \textit{ii}) PULP-Shield plugged but never turned on, to evaluate the lifetime reduction due to the additional weight introduced; \textit{iii}) PULP-Shield turned on and executing DroNet at VDD@\SI{1.0}{\volt}, FC@\SI{50}{\mega\hertz}, CL@\SI{100}{\mega\hertz}.
Our results are summarized in Table~\ref{tab:lifetime}, where as expected the biggest reduction in the lifetime is given by the increased weight.
Ultimately, the price for our visual navigation engine is $\sim22\%$ of the original lifetime.
This lifetime reduction can be curtailed through a number of optimization.
Starting from the straightforward redesign of the PCB and camera holder with lighter plastic materials (e.g., \textit{flexible substrate}), it is possible to integrate the entire electronics of the drone.
In this last case, we could either integrate the existing MCUs with the PULP SoC in the same PCB/frame or envision a PULP-based nano-drone, where the host MCU would be replaced by the PULP SoC, scheduling all the control tasks on the FC.

\subsection{Control Evaluation}

The figures of merit of this control accuracy evaluation are \textit{i}) the \textit{longest indoor traveled distance} the nano-drone is able to cover autonomously before stopping and \textit{ii}) its capability of \textit{collision avoidance} in presence of unexpected dynamic obstacles when flying at high speed.
In all the following experiments we use the most energy-efficient configuration of our visual navigation engine of VDD@\SI{1.0}{\volt}, FC@\SI{50}{\mega\hertz}, CL@\SI{100}{\mega\hertz}.
Note that all the control-loops and state estimation parameters running on the nano-drone are kept as they come with the official firmware\footnote{https://github.com/bitcraze/crazyflie-firmware}, leaving room for further improvements.

\begin{figure}[htbp]
\centerline{\includegraphics[width=\columnwidth]{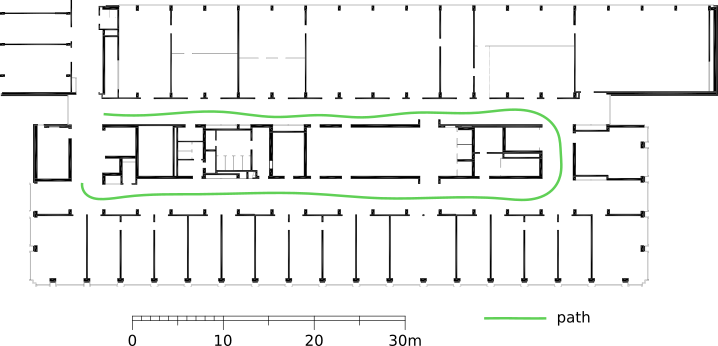}}
\caption{Testing environment: indoor corridor.}
\label{fig:indoorpath}
\end{figure}

The first control accuracy experiment is conceived to assess the capability of our nano-UAV of autonomous navigation in a previously unseen indoor environment, particularly challenging due to the visual differences from training samples (dominantly outdoor).
As shown in Figure~\ref{fig:indoorpath}, our visual navigation engine enables an indoor traveled distance of $\sim$\SI{113}{\meter}, flying at a average speed of \SI{0.5}{\meter/\second}.
The path navigated is composed of two straight corridors ($\sim$\SI{50}{\meter} each), divided by two sharp 90\textdegree~turns, resulting in the central ``U'' turn.
As shown in the supplementary video material (\youtubelink), the straight corridors are traveled with minimal modification of the indoor environment, in contrast to the ``U'' turn that requires some more auxiliary white tape on the ground to enforce the correct understanding of the surrounding by the CNN.
The flight terminates due to the glossy paint at the end of the corridor, due to the interference of the light reflection with the CNN understanding, resulting in a constant high probability of collision.

In the second part of the control accuracy evaluation, we analyze the system's capability of preventing collisions.
Correctly identifying obstacles has been already implicitly demonstrated with the autonomous navigation test.
With the collision avoidance set of experiments we want to push our visual navigation engine to its limit, preventing collisions also under very unfavourable conditions -- i.e., high flight speed and small reaction space/time.
The setup of this experiment is represented by a straight path where, after the nano-drone has traveled the first \SI{8}{\meter} at full speed, an unexpected dynamic obstacle appear within only \SI{2}{\meter} of distance from it (i.e., at \SI{10}{\meter} from the starting point).
We performed multiple experiments, sweeping the flight speed, to identify the maximum one for which the nano-drone is still able to react promptly and prevent the collision.
Results, also shown in the supplementary video material (\youtubelink), demonstrate that our visual navigation engine enables safe flight up to $\sim$\SI{1.5}{\meter/\second}.

\begin{figure}[htbp]
\centerline{\includegraphics[width=\columnwidth]{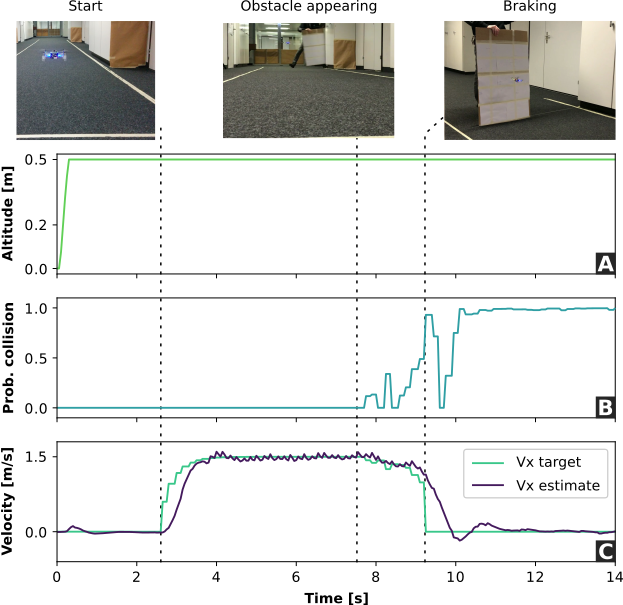}}
\caption{Onboard real-time log of the collision avoidance experiment, paired with external events.}
\label{fig:log}
\end{figure}

Figure~\ref{fig:log} reports the real-time log of the relevant onboard information (i.e., probability of collision, estimated and desired velocity, and altitude), paired with external events (i.e., start, appearing of the obstacle, and braking), of this experiment.
The initial take-off is followed by $\sim$\SI{2}{\second} of in place hovering before the output of the CNN is used and the flight in the forward direction starts. 
The altitude of this experiment is kept constant at \SI{0.5}{\meter}, as reported in Figure~\ref{fig:log}-A.
As soon as the probability of collision output from the CNN, shown in Figure~\ref{fig:log}-B, is higher of the \textit{critical probability of collision} threshold of 0.7, the target forward velocity is pushed to 0, resulting in a prompt obstacle avoidance mechanism.
The onboard state estimation of the current forward velocity (i.e., \textit{Vx estimated}) is reported in Figure~\ref{fig:log}-C paired with the desired velocity in the same direction, that is calculated on the basis of the probability of collision and bounded to the maximum forward velocity, i.e., \SI{1.5}{\meter/\second}.
If we would relax the experiment's constraints -- e.g., increasing the braking space/time -- we could enable a safe flight, avoiding collision, also at higher flight speed. 

\section{Conclusions}\label{sec:conclusion}

Nano- and pico-sized UAVs are ideal ubiquitous nodes; due to their size and physical footprint, they can act as mobile sensor hubs and data collectors for tasks such as surveillance, inspection, etc.
However, to be able to perform these tasks, they must be capable of autonomous navigation of complex environments such as the indoor of buildings and offices.
In this work, we introduce the first vertically integrated visual navigation engine for autonomous nano-UAVs field-tested in closed loop demonstrations, as shown in supplementary video materials.
Our engine consumes \SIrange{64}{272}{\milli\watt} while running at \SIrange{6}{18}{fps}, enough \textit{i)} to enable autonomous navigation on a \SI{>100}{\meter} previously unseen indoor environment, and \textit{ii)} to ensure robustness against the appearance sudden obstacles at \SI{2}{\meter} distance while flying at \SI{1.5}{\meter/\second}.
To pave the way for a huge number of advanced use-cases of autonomous nano-UAVs as wireless mobile smart sensors, we release open-source our PULP-Shield design and all code running on it, as well as datasets and trained networks.

\section*{Acknowledgment}
The authors thank Frank K. G\"urkaynak for his contribution in making the supplementary videos.
This work has been partially funded by projects EC H2020 OPRECOMP (732631) and ALOHA (780788).

\bibliographystyle{IEEEtran}
\bibliography{biblio}

\begin{thebibliography}{10}
\providecommand{\url}[1]{#1}
\csname url@samestyle\endcsname
\providecommand{\newblock}{\relax}
\providecommand{\bibinfo}[2]{#2}
\providecommand{\BIBentrySTDinterwordspacing}{\spaceskip=0pt\relax}
\providecommand{\BIBentryALTinterwordstretchfactor}{4}
\providecommand{\BIBentryALTinterwordspacing}{\spaceskip=\fontdimen2\font plus
\BIBentryALTinterwordstretchfactor\fontdimen3\font minus
  \fontdimen4\font\relax}
\providecommand{\BIBforeignlanguage}[2]{{%
\expandafter\ifx\csname l@#1\endcsname\relax
\typeout{** WARNING: IEEEtran.bst: No hyphenation pattern has been}%
\typeout{** loaded for the language `#1'. Using the pattern for}%
\typeout{** the default language instead.}%
\else
\language=\csname l@#1\endcsname
\fi
#2}}
\providecommand{\BIBdecl}{\relax}
\BIBdecl

\bibitem{dunkley14iros}
O.~Dunkley, J.~Engel, J.~Sturm, and D.~Cremers, ``Visual-inertial navigation
  for a camera-equipped 25g nano-quadrotor,'' in \emph{IROS2014 Aerial Open
  Source Robotics Workshop}, 2014.

\bibitem{40gSTM32OpticFlow}
K.~McGuire, G.~de~Croon, C.~D. Wagter, K.~Tuyls, and H.~Kappen, ``Efficient
  optical flow and stereo vision for velocity estimation and obstacle avoidance
  on an autonomous pocket drone,'' \emph{IEEE Robotics and Automation Letters},
  vol.~2, no.~2, April 2017.

\bibitem{palossi2017ultra}
D.~Palossi, A.~Marongiu, and L.~Benini, ``Ultra low-power visual odometry for
  nano-scale unmanned aerial vehicles,'' in \emph{Design, Automation \& Test in
  Europe Conference \& Exhibition (DATE), 2017}.\hskip 1em plus 0.5em minus
  0.4em\relax IEEE, 2017, pp. 1647--1650.

\bibitem{briod2013optic}
A.~Briod, J.-C. Zufferey, and D.~Floreano, ``Optic-flow based control of a 46g
  quadrotor,'' in \emph{Workshop on Vision-based Closed-Loop Control and
  Navigation of Micro Helicopters in GPS-denied Environments, IROS 2013}, no.
  EPFL-CONF-189879, 2013.

\bibitem{7080923}
X.~Zhang, B.~Xian, B.~Zhao, and Y.~Zhang, ``Autonomous flight control of a nano
  quadrotor helicopter in a gps-denied environment using on-board vision,''
  \emph{IEEE Transactions on Industrial Electronics}, vol.~62, no.~10, Oct
  2015.

\bibitem{palossi2017target}
D.~Palossi, J.~Singh, M.~Magno, and L.~Benini, ``Target following on nano-scale
  unmanned aerial vehicles,'' in \emph{2017 7th IEEE international workshop on
  advances in sensors and interfaces (IWASI)}.\hskip 1em plus 0.5em minus
  0.4em\relax IEEE, 2017, pp. 170--175.

\bibitem{kang2019generalization}
K.~Kang, S.~Belkhale, G.~Kahn, P.~Abbeel, and S.~Levine, ``Generalization
  through simulation: Integrating simulated and real data into deep
  reinforcement learning for vision-based autonomous flight,'' \emph{arXiv
  preprint arXiv:1902.03701}, 2019.

\bibitem{7487496}
K.~McGuire, G.~de~Croon, C.~de~Wagter, B.~Remes, K.~Tuyls, and H.~Kappen,
  ``Local histogram matching for efficient optical flow computation applied to
  velocity estimation on pocket drones,'' in \emph{2016 IEEE International
  Conference on Robotics and Automation (ICRA)}, May 2016.

\bibitem{dronet}
A.~Loquercio, A.~I. Maqueda, C.~R. del Blanco, and D.~Scaramuzza, ``Dronet:
  Learning to fly by driving,'' \emph{IEEE Robotics and Automation Letters},
  vol.~3, no.~2, April 2018.

\bibitem{iotUAV_survey}
N.~H. Motlagh, T.~Taleb, and O.~Arouk, ``Low-altitude unmanned aerial
  vehicles-based internet of things services: Comprehensive survey and future
  perspectives,'' \emph{IEEE Internet of Things Journal}, vol.~3, no.~6, Dec
  2016.

\bibitem{Lin_2017}
Y.~Lin, F.~Gao, T.~Qin, W.~Gao, T.~Liu, W.~Wu, Z.~Yang, and S.~Shen,
  ``Autonomous aerial navigation using monocular visual-inertial fusion,''
  \emph{Journal of Field Robotics}, vol.~35, no.~1, pp. 23--51, jul 2017.

\bibitem{loianno2018special}
G.~Loianno, D.~Scaramuzza, and V.~Kumar, ``Special issue on high-speed
  vision-based autonomous navigation of uavs,'' \emph{Journal of Field
  Robotics}, vol.~35, no.~1, pp. 3--4, 2018.

\bibitem{Wood2017}
R.~J. Wood, B.~Finio, M.~Karpelson, K.~Ma, N.~O. P{\'e}rez-Arancibia, P.~S.
  Sreetharan, H.~Tanaka, and J.~P. Whitney, \emph{Progress on ``Pico'' Air
  Vehicles}.\hskip 1em plus 0.5em minus 0.4em\relax Cham: Springer
  International Publishing, 2017.

\bibitem{conroy2009}
J.~Conroy, G.~Gremillion, B.~Ranganathan, and J.~S. Humbert, ``Implementation
  of wide-field integration of optic flow for autonomous quadrotor
  navigation,'' \emph{Autonomous robots}, vol.~27, no.~3, 2009.

\bibitem{giusti2016machine}
A.~Giusti, J.~Guzzi, D.~C. Cire{\c{s}}an, F.-L. He, J.~P. Rodr{\'\i}guez,
  F.~Fontana, M.~Faessler, C.~Forster, J.~Schmidhuber, G.~Di~Caro
  \emph{et~al.}, ``A machine learning approach to visual perception of forest
  trails for mobile robots,'' \emph{IEEE Robotics and Automation Letters},
  vol.~1, no.~2, pp. 661--667, 2016.

\bibitem{navion}
A.~Suleiman, Z.~Zhang, L.~Carlone, S.~Karaman, and V.~Sze, ``Navion: A fully
  integrated energy-efficient visual-inertial odometry accelerator for
  autonomous navigation of nano drones,'' in \emph{2018 IEEE Symposium on VLSI
  Circuits}, June 2018, pp. 133--134.

\bibitem{CNN-SLAM}
Z.~{Li}, Y.~{Chen}, L.~{Gong}, L.~{Liu}, D.~{Sylvester}, D.~{Blaauw}, and
  H.~{Kim}, ``An 879gops 243mw 80fps vga fully visual cnn-slam processor for
  wide-range autonomous exploration,'' in \emph{2019 IEEE International Solid-
  State Circuits Conference - (ISSCC)}, Feb 2019, pp. 134--136.

\bibitem{he2016deep}
K.~He, X.~Zhang, S.~Ren, and J.~Sun, ``Deep residual learning for image
  recognition,'' in \emph{Proceedings of the IEEE conference on computer vision
  and pattern recognition}, 2016, pp. 770--778.

\bibitem{contiDATE16}
F.~Conti, D.~Palossi, A.~Marongiu, D.~Rossi, and L.~Benini, ``Enabling the
  heterogeneous accelerator model on ultra-low power microcontroller
  platforms,'' in \emph{2016 Design, Automation Test in Europe Conference
  Exhibition (DATE)}, March 2016.

\bibitem{Razavian2014transfer}
A.~S. Razavian, H.~Azizpour, J.~Sullivan, and S.~Carlsson, ``{CNN} features
  off-the-shelf: An astounding baseline for recognition,'' in \emph{{IEEE}
  Conference on Computer Vision and Pattern Recognition Workshops ({CVPRW})},
  jun 2014.

\bibitem{HubaraQuantizedNeuralNetworks2016}
I.~Hubara, M.~Courbariaux, D.~Soudry, R.~{El-Yaniv}, and Y.~Bengio, ``Quantized
  {{Neural Networks}}: {{Training Neural Networks}} with {{Low Precision
  Weights}} and {{Activations}},'' \emph{arXiv:1609.07061 [cs]}, Sep. 2016.

\bibitem{xu2017end}
H.~Xu, Y.~Gao, F.~Yu, and T.~Darrell, ``End-to-end learning of driving models
  from large-scale video datasets,'' in \emph{Proceedings of the IEEE
  conference on computer vision and pattern recognition}, 2017, pp. 2174--2182.

\end{thebibliography}

\end{document}